\documentclass[conference]{IEEEtran}
\usepackage[utf8]{inputenc}
\usepackage{graphicx}
\usepackage{amsmath}
\usepackage{amsfonts}
\usepackage{amssymb}
\usepackage{cite}
\usepackage{booktabs}
\usepackage{multirow}
\usepackage{algorithm}
\usepackage{algpseudocode}
\usepackage{array}
\usepackage{hyperref}
\usepackage{caption}
\usepackage{xcolor}
\usepackage{array}
\usepackage{makecell}

\begin{document}

\title{Data-Driven Telecom Marketing Optimization: A Machine Learning-Based Churn Prediction and Customer Segmentation Framework}

\author{
    \IEEEauthorblockN{Nada Ali, Lina Ahmed}
    \IEEEauthorblockA{
        Department of Electrical and Electronic Engineering \\
        University of Khartoum \\
        \texttt{lina.alnoosh2@gmail.com}\\ \texttt{nadaali1232000@gmail.com}\\ 
    }
    \and
    \IEEEauthorblockN{Dr. Tahani Abdalla Attia}
    \IEEEauthorblockA{
        Associate Professor \\
        Department of Electrical and Electronic Engineering \\
        University of Khartoum \\
        \texttt{tahani@uofk.edu}
    }
}

\maketitle

\begin{abstract}
Customer churn is one of the most significant challenges facing telecommunication companies, directly eroding revenue and long-term customer relationships. Traditional retention programs typically rely on generic, non-personalized incentives and lack the precision needed to identify high-risk customers before they leave. This paper presents a complete, data-driven marketing optimization framework that integrates (i) machine-learning-based churn prediction, (ii) two-dimensional customer segmentation combining churn risk with customer value, and (iii) tailored, segment-specific marketing and Return-on-Investment (ROI) strategies. Using the IBM Telco Customer Churn dataset (7{,}043 customers, 21 features), three gradient-boosting ensembles---XGBoost, LightGBM, and CatBoost---were trained and rigorously tuned via randomized search with stratified 5-fold cross-validation, explicit class-weighting, and F1-score-driven decision-threshold optimization to counter the dataset's 73.4\%/26.6\% class imbalance. CatBoost was selected as the deployment model, achieving 77.68\% accuracy, 73.53\% recall, 56.12\% precision, an F1-score of 0.6366, a PR-AUC of 0.6553, and a ROC-AUC of 0.8403 on the held-out test set. Customers were subsequently partitioned with K-Means clustering (validated via the Elbow method and visualized with Principal Component Analysis) into High-, Medium-, and Low-Value segments, which were further cross-tabulated against churn-risk labels to define four actionable customer clusters. Segment-specific retention, upsell, and engagement strategies were designed for each cluster, and a theoretical ROI/CLV framework is provided to quantify the financial impact of the proposed interventions. The entire pipeline was operationalized in an interactive Streamlit web application that allows non-technical marketing teams to upload data, filter by segment, visualize churn drivers via SHAP, and download automated segment reports. The results confirm that combining predictive churn modeling with value-aware segmentation yields substantially more actionable, and more profitable, marketing decisions than churn prediction alone.
\end{abstract}


\section{Introduction}

\subsection{Background and Motivation}
Telecommunication companies operate in a saturated, price-competitive market in which acquiring a new customer is estimated to cost several times more than retaining an existing one. Customer churn---the voluntary or involuntary termination of a subscriber relationship---therefore represents a direct and compounding threat to revenue. Traditional retention programs generally apply blanket incentives (discounts, bundle offers) irrespective of a customer's actual risk of leaving or their financial value to the company, which wastes marketing budget on customers who were never going to leave while under-serving high-value customers who are quietly at risk. This paper addresses that inefficiency by building a full pipeline that predicts \emph{who} is at risk, segments \emph{how valuable} each customer is, and only then decides \emph{what} marketing action to take and \emph{whether it pays off}.

\subsection{Problem Statement}
Existing telecom retention systems suffer from three compounding weaknesses: (1) churn prediction is frequently treated as a stand-alone classification exercise, disconnected from any downstream business action; (2) customer segmentation, where it exists, is usually performed independently of churn risk, using purely descriptive demographic or usage groupings; and (3) few published systems attempt to translate model outputs into an operational, deployable tool that estimates the financial return of an intervention. This work is designed specifically to close all three gaps simultaneously within a single, reproducible pipeline.

\subsection{Contributions}
This paper makes the following contributions:
\begin{itemize}
    \item \textbf{An integrated ML pipeline.} A four-stage framework -- churn prediction, value segmentation, tailored campaign design, and ROI measurement -- implemented end-to-end on a real public dataset (Fig.~\ref{fig:overview}).
    \item \textbf{A rigorous churn-prediction study.} Three gradient boosting families (XGBoost, LightGBM, CatBoost) are trained under identical randomized-search hyperparameter optimization with 5-fold stratified cross-validation, explicit \texttt{scale\_pos\_weight} tuning, and a custom F1-maximizing threshold search (Algorithm~\ref{alg:threshold}), rather than relying on the default 0.5 decision boundary.
    \item \textbf{Two-dimensional, actionable segmentation.} K-Means clustering (with PCA-based dimensionality reduction and validation) is combined post-hoc with churn-risk labels to produce four cross-sectional customer clusters, each with a distinct, empirically profiled marketing strategy.
    \item \textbf{A transparent ROI/CLV framework.} A discounted Customer Lifetime Value formulation and campaign-level ROI equation are provided, together with worked hypothetical scenarios (Table~\ref{tab:roi_scenarios}), so telecom operators can substitute their own Average Revenue Per User (ARPU) and campaign-cost figures.
    \item \textbf{A deployable system.} The complete pipeline is implemented as an interactive Streamlit dashboard, including automated SHAP-based explainability, dynamic filtering, and downloadable per-segment text reports (Algorithm~\ref{alg:strategy}).
\end{itemize}

\subsection{Paper Structure}
Section II reviews related work in churn prediction, segmentation, and ROI-driven telecom marketing. Section III provides the theoretical background for the models used. Section IV details the methodology: dataset, preprocessing, feature engineering, model training, segmentation and the marketing/ROI framework. Section V describes the system deployment. Section VI presents experimental results. Section VII discusses key insights, addressed research gaps, and limitations. Section VIII concludes and outlines future work.

\section{Related Work}

\subsection{Churn Prediction in the Telecom Sector}
Customer churn is one of the most pressing challenges for telecom companies because of its direct impact on revenue, Customer Acquisition Cost (CAC), and long-term profitability \cite{ref1,ref14}. Numerous studies confirm that retaining an existing customer is substantially cheaper than acquiring a new one \cite{ref3,ref14}. Traditional models such as logistic regression and decision trees offer interpretability but degrade on complex, high-dimensional tabular data \cite{ref29}. Gradient boosting methods -- XGBoost \cite{ref26}, LightGBM \cite{ref27}, and CatBoost \cite{ref28} -- have since become the standard for structured churn data due to their ability to model non-linear interactions and handle mixed categorical/numerical features. Ahmad et al. applied ML on a big-data platform to forecast churn with high precision \cite{ref16}; Wagh et al. benchmarked multiple ML models and highlighted the practical value of ensembling in telecom \cite{ref15}. SyriaTel's deployment of churn prediction combined with Social Network Analysis achieved a 47\% retention rate among predicted churners \cite{ref16}, while Vodafone Ghana used survival analysis on top-up and Minutes-of-Use behavior to target at-risk high-value customers \cite{ref19}. However, most of these studies stop at the prediction step without integrating the outputs into a concrete marketing decision framework -- the gap this paper directly targets.

\subsection{ML in Marketing Optimization}
Telecom marketing has shifted from mass promotion toward personalized, data-driven campaigns. ML enables near-real-time segmentation, targeting, and campaign adjustment to maximize ROI and Customer Lifetime Value (CLV) \cite{ref5,ref12,ref24}. Integrating ML outputs with CRM systems allows promotions to be tailored to behavior, preference, and churn risk \cite{ref6,ref25}. Sridevi et al. proposed predictive analytics for end-to-end campaign optimization, including offer generation and ROI tracking \cite{ref12}. Despite this progress, few works propose a full-stack implementation that unifies prediction, segmentation, and personalized marketing in a single deployable system -- a gap this study addresses directly with an operational Streamlit application.

\subsection{Customer Segmentation Techniques}
Traditional segmentation relies on demographic or usage-based groupings, which often lack the resolution needed to identify nuanced behavior or profitability potential \cite{ref21,ref22}. Modern approaches use K-Means, hierarchical clustering, and dimensionality-reduction methods such as PCA to generate higher-resolution segments \cite{ref20,ref23}, distinguishing groups such as high-value loyalists, price-sensitive users, or dormant subscribers. Alkhayrat et al. compared deep-learning-based dimensionality reduction against classical PCA for telecom clustering \cite{ref20}; Sudirjo emphasized how data-driven segmentation accelerates marketing decision-making \cite{ref22}. Segmentation efforts, however, are rarely integrated with churn-risk labels or ROI-focused campaign planning -- a limitation this work overcomes by cross-tabulating value segments directly with churn-risk output.

\subsection{ROI and CLV in Telecom Marketing}
ROI and CLV are the key financial metrics guiding telecom marketing decisions. Ramachandran defines CLV as the present value of all future profits a customer will generate, essential for prioritizing retention and upsell spend \cite{ref5}. Sridevi et al. proposed predictive analytics to track ROI alongside customer behavior \cite{ref12}, and Pandian argued for evaluating retention in terms of measurable financial return rather than generic engagement metrics \cite{ref4}. The operational challenge of combining these financial metrics with real-time ML output in a single tool remains open; Section~\ref{sec:roi} proposes exactly such a unified, adjustable framework.

\section{Theoretical Background}

\subsection{Gradient Boosting Framework}
Gradient boosting builds an additive ensemble of weak learners (decision trees) $h_m(x)$, each fitted sequentially to the residual/negative-gradient of a differentiable loss function:
\begin{equation}
    \hat{y} = \sum_{m=1}^{M} \gamma_m h_m(x)
    \label{eq:boosting}
\end{equation}
where $\gamma_m$ is the weight assigned to the $m$-th learner. For binary churn classification, the model minimizes a regularized logistic-loss objective:
\begin{equation}
    Obj(\theta) = \sum_{i=1}^{n} \ell(y_i, \hat{y}_i) + \sum_{t=1}^{T} \Omega(f_t)
    \label{eq:obj}
\end{equation}
where $\ell(y_i, \hat{y}_i) = -\left[y_i \log(\hat{y}_i) + (1-y_i)\log(1-\hat{y}_i)\right]$ is the logistic loss and $\Omega(f_t)$ is a regularization term penalizing tree complexity. All three algorithms studied here minimize \eqref{eq:obj}, but differ substantially in tree-growth strategy, categorical-feature handling and regularization, as summarized in Table~\ref{tab:model_comparison}.

\begin{table}[t]
\centering
\caption{Comparison of the Three Gradient-Boosting Churn Models}
\label{tab:model_comparison}
\begin{tabular}{|p{1.6cm}|p{1.9cm}|p{1.9cm}|p{1.9cm}|}
\hline
\textbf{Aspect} & \textbf{XGBoost} & \textbf{LightGBM} & \textbf{CatBoost} \\
\hline
Tree growth & Level-wise (breadth-first) & Leaf-wise (best-first) & Oblivious (symmetric splits) \\
\hline
Categorical handling & Requires explicit encoding & Requires explicit encoding & Native \texttt{cat\_features} support \\
\hline
Regularization & $L_1$ (\texttt{reg\_alpha}), $L_2$ (\texttt{reg\_lambda}) & $L_1$, $L_2$ & $L_2$ (\texttt{l2\_leaf\_reg}) + ordered boosting \\
\hline
Overfitting control & \texttt{gamma}, \texttt{min\_child} \texttt{\_weight}, pruning & \texttt{min\_data\_in} \texttt{\_leaf}, \texttt{num\_leaves} & \texttt{depth}, permutation-driven ordered boosting \\
\hline
Parallelism & Data / boosting-round parallel & Exclusive feature bundling & Native CPU/GPU optimization \\
\hline
\end{tabular}
\end{table}

\subsection{K-Means Clustering}
K-Means partitions customers into $K$ clusters by minimizing the within-cluster sum of squares (WCSS):
\begin{equation}
    \text{WCSS} = \sum_{k=1}^{K} \sum_{x_i \in C_k} \lVert x_i - \mu_k \rVert^2
    \label{eq:wcss}
\end{equation}
where $C_k$ is the $k$-th cluster and $\mu_k$ its centroid. Centroids are initialized with the $K$-means++ scheme and updated iteratively until convergence. The optimal $K$ was chosen using the Elbow method (Fig.~\ref{fig:elbow}), which plots \eqref{eq:wcss} against $K$ and identifies the point of diminishing returns; $K=3$ was selected in this study.

\subsection{Principal Component Analysis (PCA)}
PCA is used both to decorrelate clustering features and to visualize the resulting segments in two dimensions. Given the standardized feature matrix $X$, the covariance matrix is
\begin{equation}
    \text{Cov}(X) = \frac{1}{n-1}X^{\top}X
    \label{eq:cov}
\end{equation}
and the data is projected onto the leading eigenvectors $W$ of $\text{Cov}(X)$:
\begin{equation}
    Z = XW
    \label{eq:pca_proj}
\end{equation}
Components were retained according to their explained-variance ratio,
\begin{equation}
    \text{EVR}_j = \frac{\lambda_j}{\sum_{p=1}^{P}\lambda_p}
    \label{eq:evr}
\end{equation}
which preserves maximal variance while eliminating redundant/noisy dimensions prior to clustering and visualization (Fig.~\ref{fig:pca_clusters}).

\subsection{Evaluation Metrics}
Given the binary nature of churn prediction and the class imbalance in the dataset, the following metrics were used:
\begin{equation}
    \text{Accuracy} = \frac{TP+TN}{TP+TN+FP+FN}
    \label{eq:acc}
\end{equation}
\begin{equation}
    \text{Precision} = \frac{TP}{TP+FP}, \qquad
    \text{Recall} = \frac{TP}{TP+FN}
    \label{eq:precrec}
\end{equation}
\begin{equation}
    F1 = 2\cdot\frac{\text{Precision}\cdot\text{Recall}}{\text{Precision}+\text{Recall}}
    \label{eq:f1}
\end{equation}
Because accuracy is misleading under a 73.4\%/26.6\% class split, ROC-AUC and, more importantly, Precision-Recall AUC (PR-AUC) were used as the primary model-selection criteria, since PR-AUC is far more sensitive to minority-class (churner) performance than ROC-AUC under imbalance.

\section{Methodology}

\begin{figure}[t]
    \centering
    \includegraphics[width=0.48\textwidth]{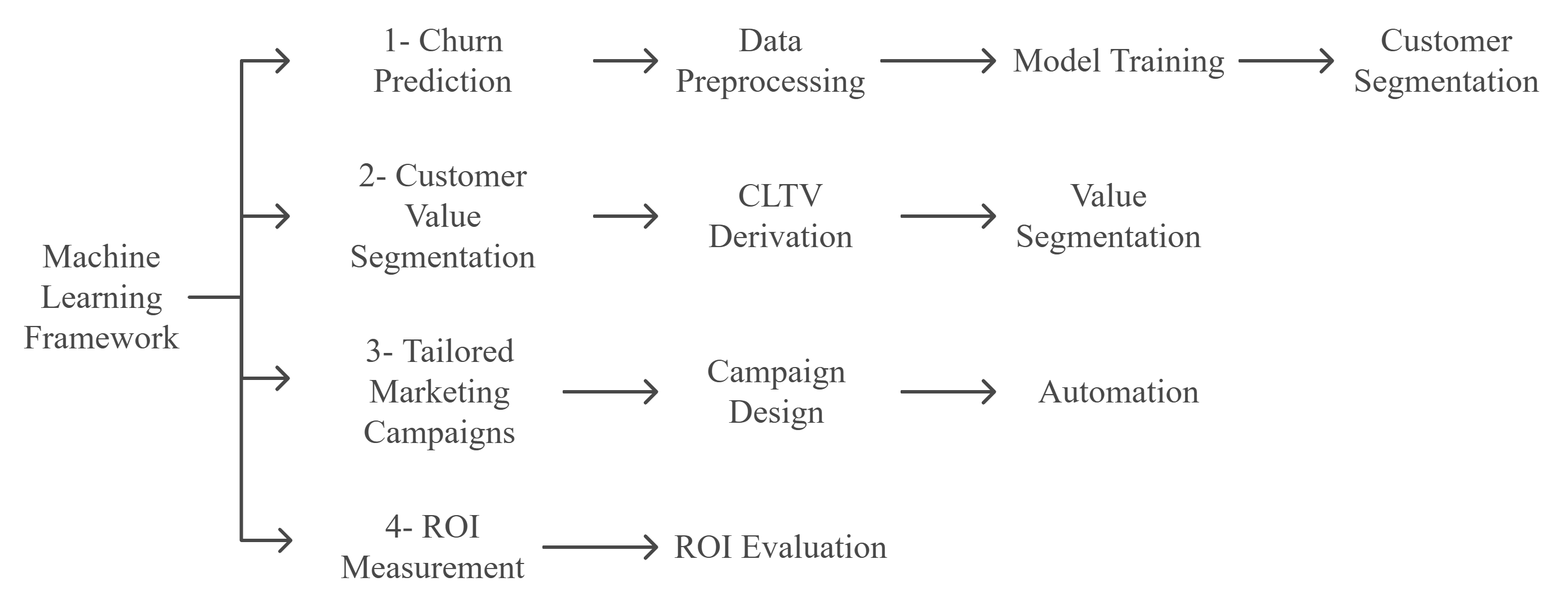}
    \caption{Overview of the proposed four-stage ML framework.}
    \label{fig:overview}
\end{figure}

Fig.~\ref{fig:overview} shows the overall four-stage framework: churn prediction $\rightarrow$ customer value segmentation $\rightarrow$ tailored marketing campaigns $\rightarrow$ ROI measurement.

\subsection{Dataset Description}
This study uses the publicly available \textbf{IBM Telco Customer Churn dataset} \cite{ref28b}, retrieved from Kaggle, comprising \textbf{7{,}043 customer records and 21 features} spanning demographics, account information, service subscriptions, and billing. The target variable, \texttt{Churn}, is binary (Yes/No). Table~\ref{tab:dataset_char} summarizes the dataset's key characteristics.

\begin{table}[t]
\centering
\caption{Characteristics of the IBM Telco Dataset}
\label{tab:dataset_char}
\begin{tabular}{|l|r|}
\hline
Total features & 21 \\
\hline
Total customers & 7{,}043 \\
\hline
Missing values & None (initially) \\
\hline
Churn class & 26.6\% (1{,}869) \\
\hline
Non-churn class & 73.4\% (5{,}174) \\
\hline
Data distribution & Imbalanced \\
\hline
Categorical features & 17 \\
\hline
Numerical features & 4 \\
\hline
\end{tabular}
\end{table}

Feature groups include: \emph{Usage/Financial} (\texttt{tenure}, \texttt{MonthlyCharges}, \texttt{TotalCharges}); \emph{Service subscription} (\texttt{InternetService}, \texttt{OnlineSecurity}, \texttt{OnlineBackup}, \texttt{DeviceProtection}, \texttt{TechSupport}, \texttt{StreamingTV}, \texttt{StreamingMovies}, \texttt{PhoneService}, \texttt{MultipleLines}); \emph{Account} (\texttt{Contract}, \texttt{PaperlessBilling}, \texttt{PaymentMethod}); and \emph{Demographic} (\texttt{gender}, \texttt{SeniorCitizen}, \texttt{Partner}, \texttt{Dependents}).

\subsubsection{Dataset Limitations}
\begin{figure}[t]
    \centering

    \includegraphics[width=0.3\textwidth]{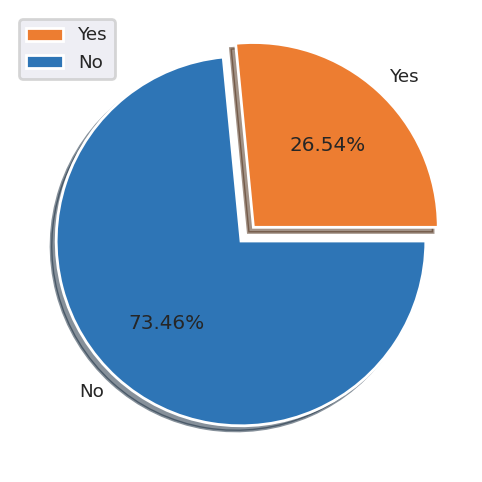}
    \caption{Class distribution of the target variable Churn.}
    \label{fig:churn_dist}
\end{figure}

Four limitations were identified and explicitly accounted for during modeling: (1) \emph{class imbalance} (5{,}174 vs. 1{,}869, Fig.~\ref{fig:churn_dist}); (2) \emph{limited temporal coverage} -- the dataset is a single snapshot with no time-series or seasonal signal; (3) \emph{no external variables} such as competitor actions or macroeconomic conditions; and (4) \emph{potential label oversimplification}, since the binary Yes/No churn label cannot capture partial or temporary service reduction.

\subsection{Exploratory Data Analysis}
\begin{figure}[t]
    \centering
    \includegraphics[width=0.48\textwidth]{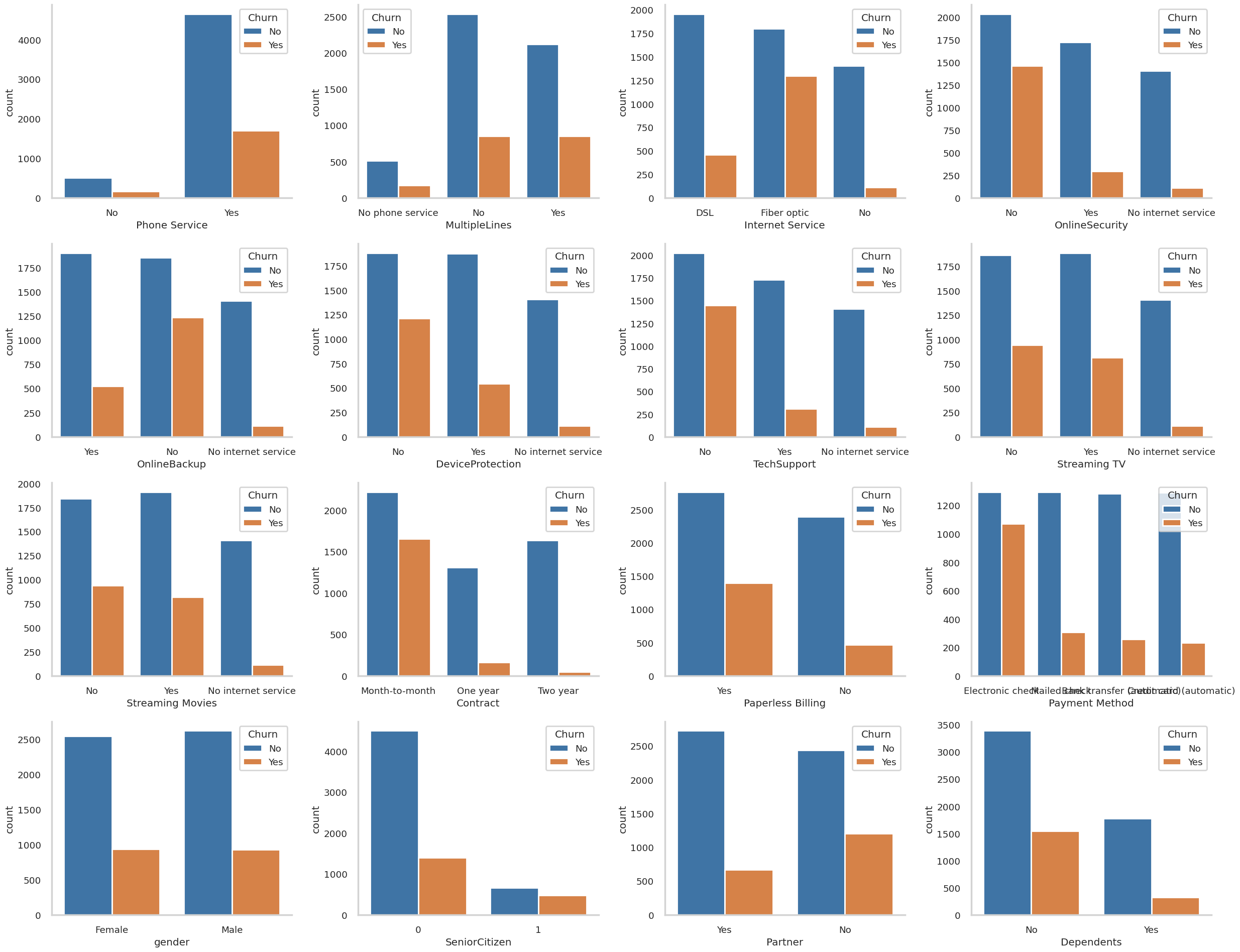}
    \caption{Churn distribution across categorical features.}
    \label{fig:cat_churn}
\end{figure}

\begin{figure}[t]
    \centering
    \includegraphics[width=0.48\textwidth]{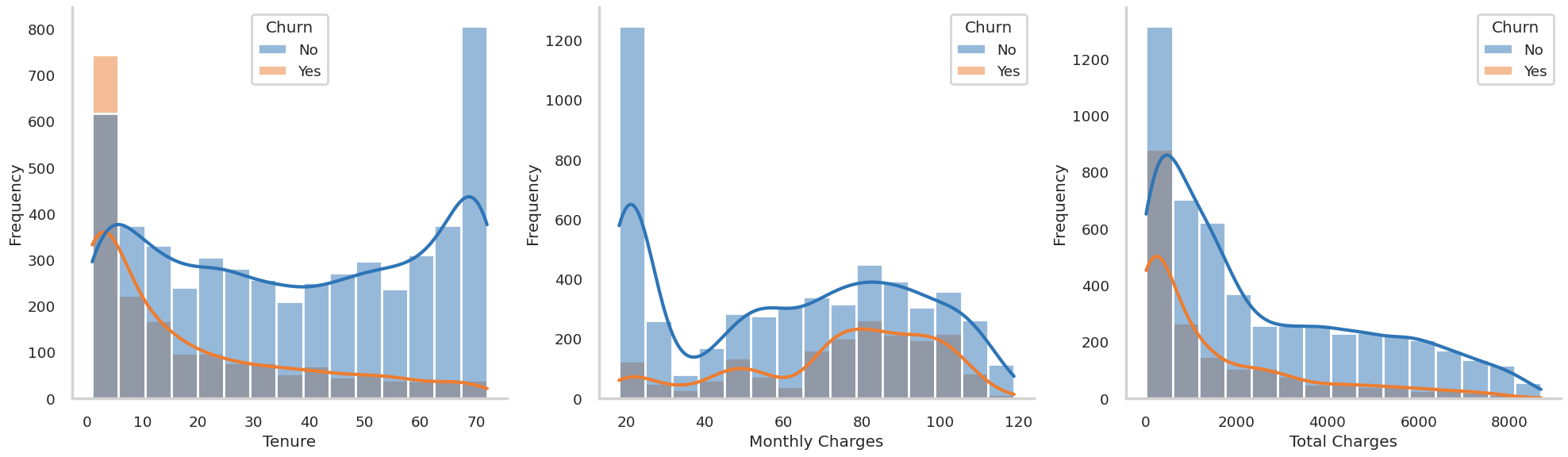}
    \caption{Distribution of numerical features by churn status.}
    \label{fig:num_churn}
\end{figure}

Bar-chart analysis of categorical features against churn (Fig.~\ref{fig:cat_churn}) shows the highest churn concentrated among month-to-month contract holders, fiber-optic internet subscribers (vs. DSL), customers lacking online security/tech support/backup add-ons, customers paying via electronic check, and senior citizens without a partner or dependents. Distribution analysis of numerical features (Fig.~\ref{fig:num_churn}) shows churn is highest among short-tenure, mid-to-high-paying customers, while long-tenure, high-total-charge customers are the most loyal -- confirming tenure and contract type as leading churn predictors.

\subsection{Data Preprocessing}
The following preprocessing pipeline was applied prior to modeling:
\begin{itemize}
    \item \textbf{Irrelevant-column removal:} \texttt{customerID} was dropped.
    \item \textbf{Missing values:} converting \texttt{TotalCharges} to numeric revealed 11 rows with missing values, all corresponding to \texttt{tenure}=0 (logically inconsistent); these rows were removed via listwise deletion.
    \item \textbf{Duplicate removal:} 22 duplicated rows were identified and dropped.
    \item \textbf{Categorical encoding:} one-hot encoding (with \texttt{drop='first'}) for XGBoost/LightGBM; CatBoost consumed raw categorical columns natively via its \texttt{cat\_features} parameter, avoiding target leakage from encoding.
    \item \textbf{Target encoding:} \texttt{Churn} mapped Yes/No $\rightarrow$ 1/0.
    \item \textbf{Stratified 80/20 train-test split} to preserve the 73.4\%/26.6\% class ratio in both partitions.
\end{itemize}

\subsection{Feature Engineering}
Fifteen domain-informed features were engineered to capture financial stress, service instability, and engagement signals not directly present in the raw schema (Table~\ref{tab:features}).

\begin{table}[t]
\centering
\caption{Newly Engineered Features}
\label{tab:features}
\footnotesize
\renewcommand{\arraystretch}{1.3}
\begin{tabular}{|>{\centering\arraybackslash}p{2.4cm}|>{\raggedright\arraybackslash}p{5.0cm}|}
\hline
\textbf{Feature} & \textbf{Purpose} \\
\hline
\texttt{\makecell[l]{Monthly\\Charges\_to\_\\Total\_Ratio}} & Short-term vs. long-term financial commitment \\
\hline
\texttt{NEW\_Engaged} & Flags 1-/2-year contracts (lower churn propensity) \\
\hline
\texttt{\makecell[l]{Lack\_of\_\\Protections}} & Missing device/tech protection (security risk) \\
\hline
\texttt{\makecell[l]{Young\_\\Unengaged\_\\Customers}} & Young customers on short-term contracts \\
\hline
\texttt{\makecell[l]{Total\_Active\_\\Services}} & Count of active services (engagement proxy) \\
\hline
\texttt{\makecell[l]{NEW\_FLAG\_\\ANY\_STREAMING}} & Any streaming subscription (modern usage) \\
\hline
\texttt{\makecell[l]{AutoPayment\_\\Usage}} & Automatic payment method (payment reliability) \\
\hline
\texttt{\makecell[l]{Average\_\\Charge\_Per\_\\Month}} & Mean monthly spend over tenure \\
\hline
\texttt{NEW\_Increase} & Current vs. average monthly charge \\
\hline
\texttt{\makecell[l]{Average\_\\Service\_Cost}} & Cost per active service (value perception) \\
\hline
\texttt{\makecell[l]{Unstable\_\\Payment\_\\Behavior}} & Manual payment + short contract (instability) \\
\hline
\texttt{\makecell[l]{HighRisk\_\\Low\_Tenure}} & New customers ($<$6 months) lacking security \\
\hline
\texttt{Tenure\_Group} & Lifecycle-stage segmentation by tenure \\
\hline
\texttt{\makecell[l]{Service\_\\Instability}} & Security + tech-support gap indicator \\
\hline
\texttt{\makecell[l]{HighSpend\_\\LowTenure}} & High spend + low tenure (retention priority) \\
\hline
\end{tabular}
\end{table}

\subsection{Churn Prediction: Training and Hyperparameter Optimization}
Each model was trained with \texttt{RandomizedSearchCV} under stratified 5-fold cross-validation, optimizing \emph{average precision} (equivalent to PR-AUC) as the search objective. The tuned hyperparameter spaces covered:
\begin{itemize}
    \item Class weighting: \texttt{scale\_pos\_weight}
    \item Learning rate: \texttt{learning\_rate}
    \item Regularization: \texttt{reg\_alpha}, \texttt{reg\_lambda}, \texttt{l2\_leaf\_reg}
    \item Tree structure/complexity: \texttt{max\_depth}, \texttt{depth}, \texttt{num\_leaves}, \texttt{min\_data\_in\_leaf}, \texttt{min\_child\_weight}, \texttt{gamma}
    \item Ensemble size: \texttt{n\_estimators}, \texttt{iterations}
\end{itemize}

\subsubsection{Class Imbalance Mitigation}
The 1:2.76 churn/non-churn ratio was addressed with two complementary strategies rather than resampling alone:
\begin{enumerate}
    \item \textbf{Dynamic class weights:} \texttt{scale\_pos\_weight} was searched using both the raw imbalance ratio (2.76) and its square root (1.66), balancing sensitivity against over-penalization of the majority class.
    \item \textbf{Threshold optimization:} rather than the default 0.5 cutoff, probabilistic predictions from 3-fold cross-validation were swept across 100 thresholds $\in [0.1, 0.9]$, selecting the threshold that maximized F1-score subject to a minimum precision and recall of 0.7 each (Algorithm~\ref{alg:threshold}).
\end{enumerate}

\begin{algorithm}[t]
\caption{F1-Maximizing Decision Threshold Search}
\label{alg:threshold}
\begin{algorithmic}[1]
\Require Cross-validated churn probabilities $\hat{p}_i$, true labels $y_i$
\Ensure Optimal decision threshold $t^{*}$
\State $T \gets \{0.10, 0.109, \dots, 0.90\}$ \Comment{100 evenly spaced thresholds}
\State $\text{best\_f1} \gets -\infty$, $t^{*} \gets 0.5$
\For{each $t \in T$}
    \State $\hat{y}_i \gets \mathbb{1}[\hat{p}_i \geq t]$
    \State compute $\text{Precision}(t)$, $\text{Recall}(t)$, $F1(t)$ via \eqref{eq:precrec}, \eqref{eq:f1}
    \If{$\text{Precision}(t) \geq 0.7$ \textbf{and} $\text{Recall}(t) \geq 0.7$}
        \If{$F1(t) > \text{best\_f1}$}
            \State $\text{best\_f1} \gets F1(t)$, $t^{*} \gets t$
        \EndIf
    \EndIf
\EndFor
\If{no threshold satisfied the constraint}
    \State $t^{*} \gets \arg\max_t F1(t)$ \Comment{unconstrained fallback}
\EndIf
\State \Return $t^{*}$
\end{algorithmic}
\end{algorithm}

\subsection{Customer Value Segmentation}
\label{sec:segmentation}
\begin{figure}[t]
    \centering
    \includegraphics[width=0.4\textwidth]{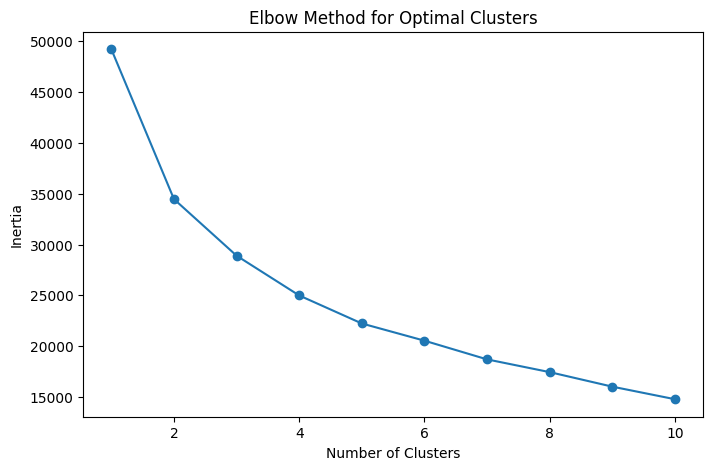}
    \caption{Elbow method confirming $K=3$ as the optimal cluster count.}
    \label{fig:elbow}
\end{figure}

\begin{figure}[t]
    \centering
    \includegraphics[width=0.45\textwidth]{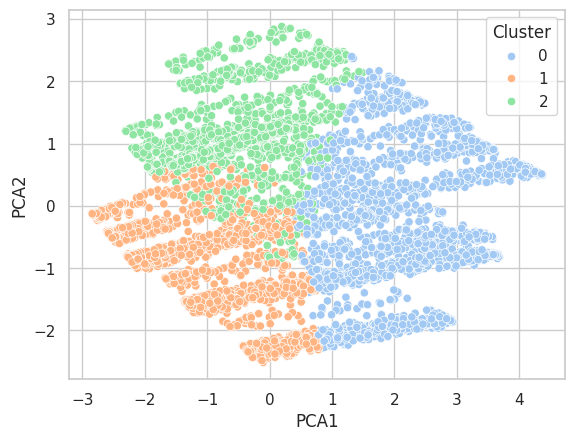}
    \caption{Customer clusters visualized on the first two principal components.}
    \label{fig:pca_clusters}
\end{figure}

Independently of churn-risk labels, customers were segmented by \emph{value} using K-Means clustering ($K=3$, selected via the Elbow method, Fig.~\ref{fig:elbow}) on standardized \texttt{tenure}, \texttt{MonthlyCharges}, \texttt{TotalCharges}, and service-adoption features (\texttt{OnlineSecurity}, \texttt{TechSupport}, \texttt{StreamingTV}, \texttt{StreamingMovies}). PCA was then applied for two-dimensional visualization (Fig.~\ref{fig:pca_clusters}) using \eqref{eq:pca_proj}.

The resulting clusters were mapped to value categories: \textbf{High-Value} (Cluster 0: long tenure, high spend, low churn risk); \textbf{Medium-Value} (Cluster 1: moderate cost, high churn risk); and \textbf{Low-Value} (Cluster 2: low cost/usage, low discontinuation risk). Cross-tabulating these three value tiers against the churn-risk label from Section IV-E produces the four operational clusters used for marketing design: \emph{High-Risk/High-Value}, \emph{Low-Risk/High-Value}, \emph{Medium-Value} (both risk levels), and \emph{Low-Risk/Low-Value}. Detailed empirical profiles for each combination -- contract mix, internet service type, payment method distribution, and service-adoption rates -- are given in Tables~\ref{tab:highvalue}--\ref{tab:lowvalue}.

\begin{table}[t]
\centering
\caption{Characteristics of High-Value Customers}
\label{tab:highvalue}
\footnotesize
\begin{tabular}{|l|p{2.1cm}|p{2.1cm}|}
\hline
\textbf{Characteristic} & \textbf{High-Risk} & \textbf{Low-Risk} \\
\hline
Total customers & 428 (6.1\%) & 1{,}656 (23.5\%) \\
\hline
Monthly charges & \$42{,}712.20 (9.4\%) & \$151{,}310.65 (33.2\%) \\
\hline
Contract mix & 92.1\% month-to-month, 7.9\% one-year & 51.3\% two-year, 37.3\% one-year, 11.4\% month-to-month \\
\hline
Internet service & 97.4\% fiber, 2.6\% DSL & 59.2\% fiber, 40.8\% DSL \\
\hline
Payment method & 65.7\% electronic check & 33.6\% bank transfer, 33.5\% credit card \\
\hline
Streaming TV/Movies & 94.9\% / 93.7\% & 81.3\% / 82.3\% \\
\hline
Tech support / Sec. & 33.4\% / 24.5\% & 65.4\% / 61.1\% \\
\hline
\end{tabular}
\end{table}

\begin{table}[t]
\centering
\caption{Characteristics of Medium-Value Customers}
\label{tab:medvalue}
\footnotesize
\begin{tabular}{|l|p{2.1cm}|p{2.1cm}|}
\hline
\textbf{Characteristic} & \textbf{Low-Risk} & \textbf{High-Risk} \\
\hline
Total customers & 450 (6.4\%) & 1{,}871 (26.6\%) \\
\hline
Avg. monthly charge & \$60.42 & \$74.05 \\
\hline
Contract mix & 60.4\% month-to-month, 34.9\% one-year & 99.8\% month-to-month \\
\hline
Internet service & 69.3\% DSL, 30.7\% fiber & 22.9\% DSL, 77.1\% fiber \\
\hline
Payment method & Balanced across all & 63.6\% electronic check \\
\hline
Streaming adoption & 27\% & 33.5\% \\
\hline
\end{tabular}
\end{table}

\begin{table}[t]
\centering
\caption{Characteristics of Low-Value Customers}
\label{tab:lowvalue}
\footnotesize
\begin{tabular}{|l|p{2.1cm}|p{2.1cm}|}
\hline
\textbf{Characteristic} & \textbf{Low-Risk} & \textbf{High-Risk} \\
\hline
Total customers & 2{,}289 (32.6\%) & 338 (4.8\%) \\
\hline
Contract mix & 35.6\% M2M, 35.6\% two-year & 100\% month-to-month \\
\hline
Internet service & 60.9\% none, 36.0\% DSL & 37.3\% none, 48.8\% DSL \\
\hline
Payment method & 39.1\% mailed check & 46.2\% mailed check, 34.0\% electronic check \\
\hline
Streaming adoption & 7--8\% & 9--10\% \\
\hline
\end{tabular}
\end{table}

\subsection{Marketing Campaign Optimization}
Marketing actions were designed per cluster: \emph{high-risk, high-value} customers receive high-touch, personalized retention (free service months, exclusive content, flexible payment plans); \emph{low-risk, high-value} customers receive scalable, cost-conscious loyalty reinforcement (automated communication, loyalty discounts, contract upgrades); \emph{medium-value} customers receive a mix of upselling (tiered incentives, bundled services) and, for the high-risk sub-group, more aggressive cost-conscious retention; and \emph{low-value} customers receive lightweight engagement tactics (small discounts, referral programs, free streaming trials, family bundles) to control cost while maintaining activity. Strategy selection is automated via a lookup table indexed by (churn-risk, customer-value) as shown in Algorithm~\ref{alg:strategy}.

\begin{algorithm}[t]
\caption{Marketing Strategy Retrieval}
\label{alg:strategy}
\begin{algorithmic}[1]
\Require Selected filters: \texttt{churn\_risk}, \texttt{customer\_value}
\Ensure List of recommended marketing strategies
\State Define \texttt{MARKETING\_STRATEGIES} as a dictionary mapping\newline
\hspace*{1em} $(\text{churn\_risk}, \text{customer\_value}) \rightarrow$ list of strategies
\State $key \gets (\texttt{churn\_risk}, \texttt{customer\_value})$
\If{$key \in$ \texttt{MARKETING\_STRATEGIES}}
    \State \Return \texttt{MARKETING\_STRATEGIES}$[key]$
\Else
    \State \Return \textit{``No specific strategy defined for this segment.''}
\EndIf
\end{algorithmic}
\end{algorithm}

\subsection{Return on Investment (ROI) Analysis Framework}
\label{sec:roi}
Because proprietary financial data was unavailable, a transparent, substitutable theoretical framework was developed instead of reporting a single unverifiable ROI figure. Estimated customer lifetime (in months) is derived from the predicted annual churn rate:
\begin{equation}
    \text{Lifetime} = \frac{12}{\text{Predicted Annual Churn Rate}}
    \label{eq:lifetime}
\end{equation}
A retention intervention with efficacy rate $e \in [0.15, 0.30]$ (a conservative literature-informed range) extends this baseline lifetime multiplicatively:
\begin{equation}
    \text{Lifetime}_{\text{ext}} = \text{Lifetime}_{\text{base}} \times (1+e)
    \label{eq:lifetime_ext}
\end{equation}
The discounted Customer Lifetime Value over horizon $T$ is
\begin{equation}
    CLV = \sum_{t=1}^{T} \frac{R_t\cdot(1-\text{Churn}_t)}{(1+d)^t}
    \label{eq:clv}
\end{equation}
where $R_t$ is revenue at month $t$, $\text{Churn}_t$ the predicted monthly churn probability, and $d$ the discount rate. Campaign-level financial return is then
\begin{equation}
    \text{ROI} = \frac{(\text{ARPU}\times\text{Lifetime}_{\text{ext}}) - \text{Campaign Cost}}{\text{Campaign Cost}}
    \label{eq:roi}
\end{equation}
Table~\ref{tab:roi_scenarios} presents hypothetical scenarios illustrating how targeting efficiency scales with segment value; operators should substitute their own ARPU and campaign-cost figures per Section~\ref{sec:limitations}.

\begin{table}[t]
\centering
\caption{Hypothetical ROI Improvement Scenarios}
\label{tab:roi_scenarios}
\begin{tabular}{|p{1.5cm}|c|c|c|}
\hline
\textbf{Segment} & \textbf{Churn Reduction} & \textbf{CLV Increase} & \textbf{ROI Multiple} \\
\hline
High-Risk, High-Value & 25\% & +35\% & 3.5$\times$ \\
\hline
High-Risk, Medium-Value & 20\% & +22\% & 2.8$\times$ \\
\hline
Low-Risk, High-Value & 10\% & +15\% & 4.1$\times$ \\
\hline
\end{tabular}
\end{table}

The framework's key insight is that targeting efficiency should be prioritized by marginal value per unit cost,
\begin{equation}
    \frac{\Delta CLV}{\text{Cost}} \rightarrow \max
    \label{eq:targeting}
\end{equation}
which explains why the low-risk/high-value segment shows the highest ROI multiple despite a smaller absolute churn reduction: the low intervention cost of loyalty-reinforcement campaigns dominates the ratio.

\section{System Deployment}

\begin{figure}[t]
    \centering
    \includegraphics[width=0.45\textwidth]{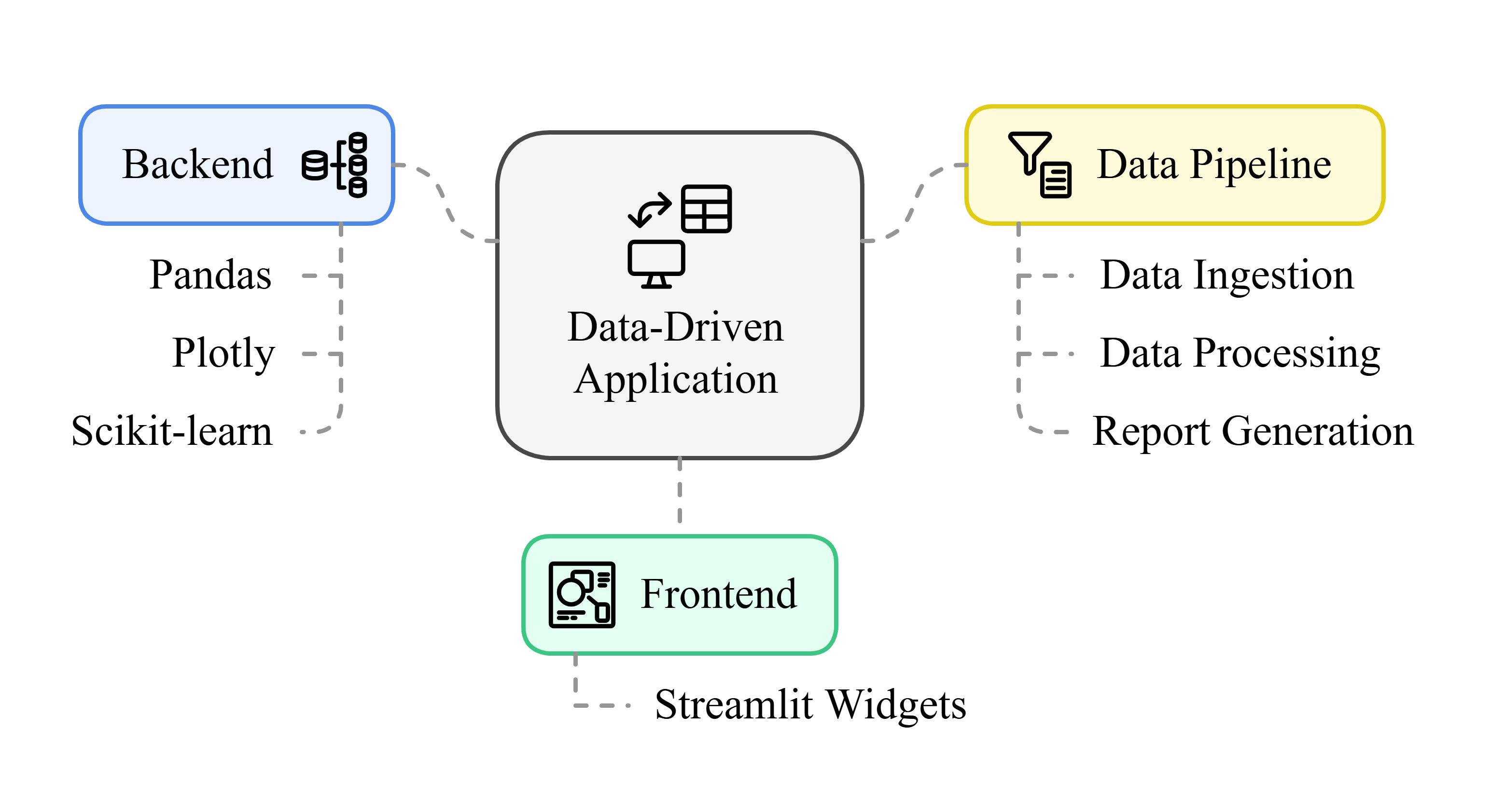}
    \caption{Three-tier system architecture: frontend (Streamlit widgets), backend (Pandas/Plotly/Scikit-learn), and data pipeline.}
    \label{fig:architecture}
\end{figure}

\begin{figure}[t]
    \centering
    \includegraphics[width=0.48\textwidth]{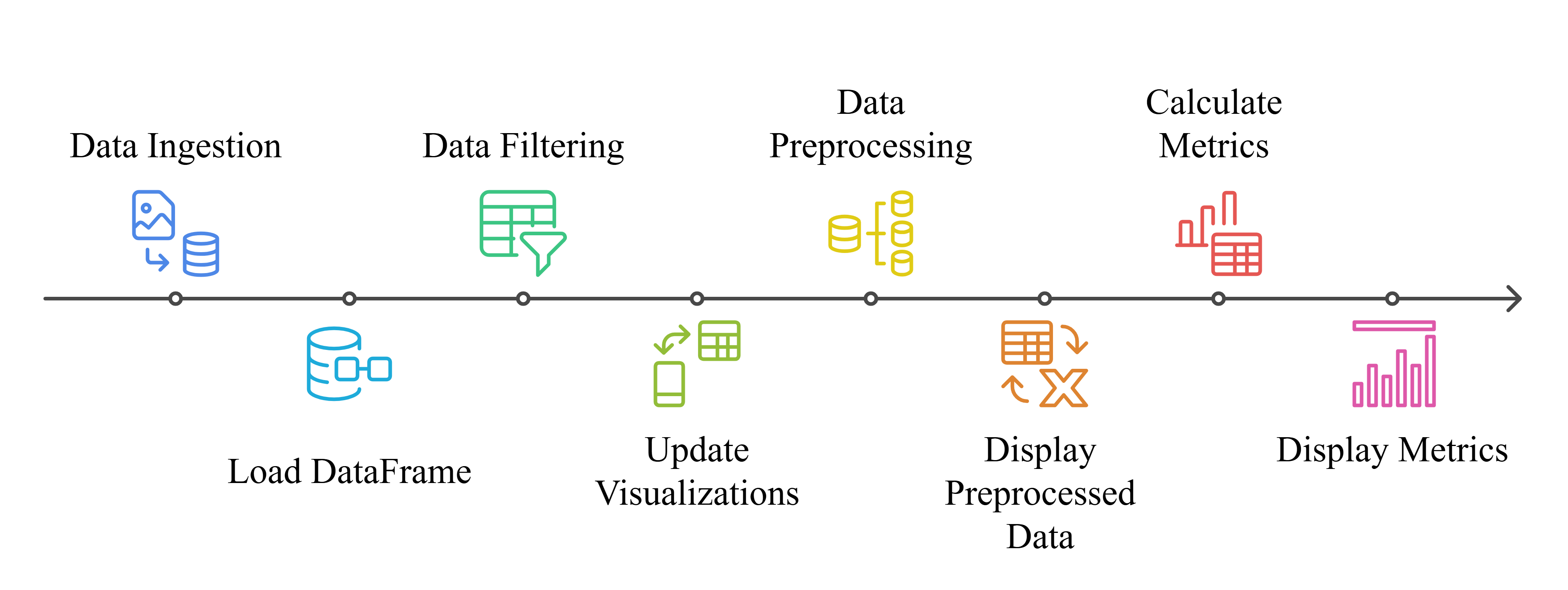}
    \caption{Customer data processing sequence, from CSV ingestion through filtering, preprocessing, metric computation, and visualization refresh.}
    \label{fig:sequence}
\end{figure}

To move beyond an offline notebook study, the complete pipeline was operationalized as an interactive web application using the \textbf{Streamlit} framework, so that a non-technical marketing analyst -- not just the model's authors -- can act on its output. The system follows a Model-View-Controller pattern adapted for data-science applications: the \emph{Model} handles preprocessing and inference (Pandas for data wrangling, Scikit-learn for encoding, the tuned CatBoost classifier for scoring), the \emph{View} is the Streamlit front end rendered with Plotly, and the \emph{Controller} manages file upload, filter state, and session state across reruns. Internally, the application executes a five-stage data pipeline -- ingestion, preprocessing, analysis, visualization, and report generation -- summarized architecturally in Fig.~\ref{fig:architecture} and traced end-to-end as a sequence diagram in Fig.~\ref{fig:sequence}.

\subsection{Landing Page and Data-Control Sidebar}
\begin{figure}[t]
    \centering
    \includegraphics[width=0.48\textwidth]{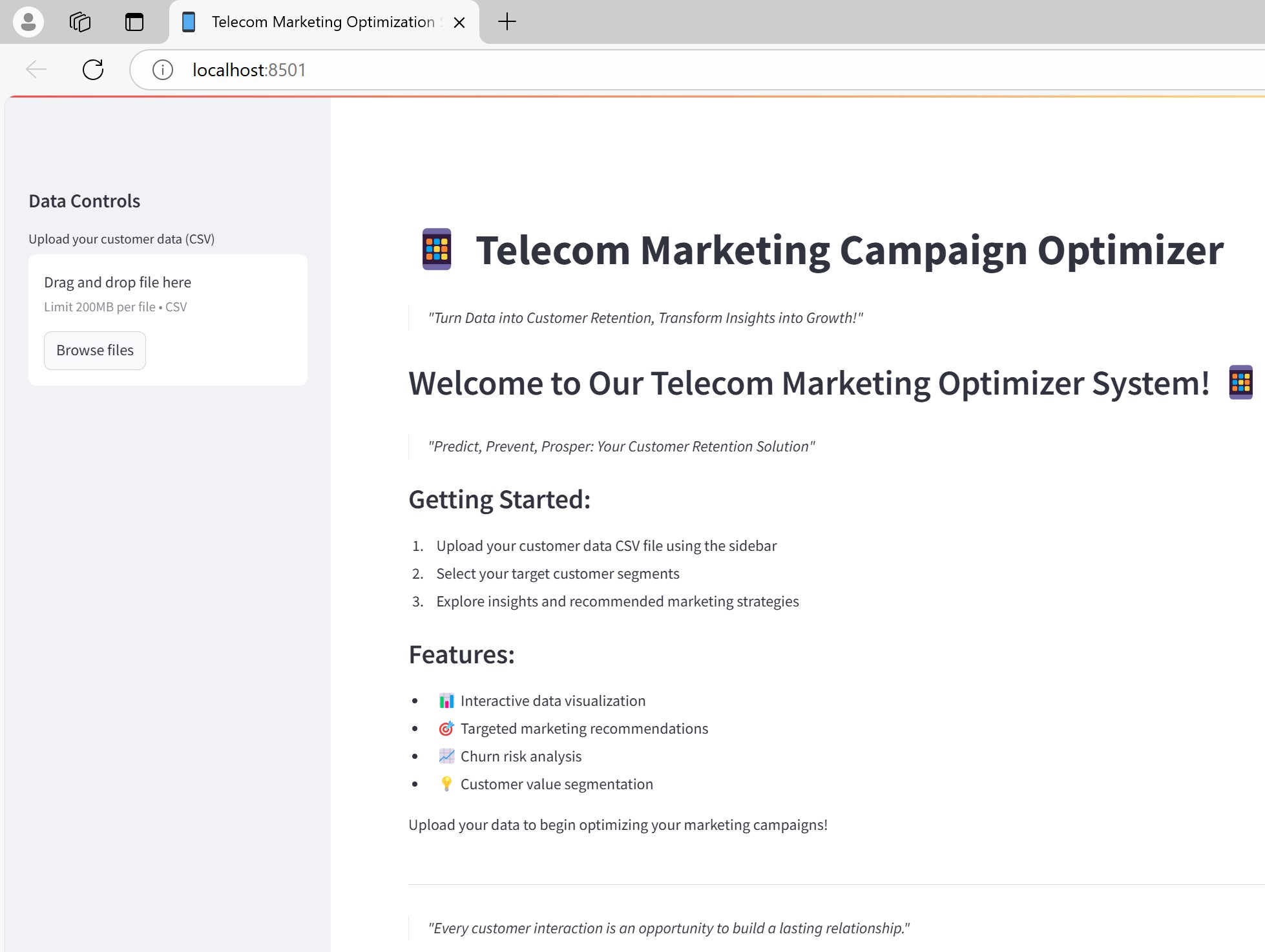}
    \caption{Landing page of the Telecom Marketing Campaign Optimizer, with onboarding instructions and a feature summary.}
    \label{fig:landing}
\end{figure}

\begin{figure}[t]
    \centering
    \includegraphics[width=0.38\textwidth]{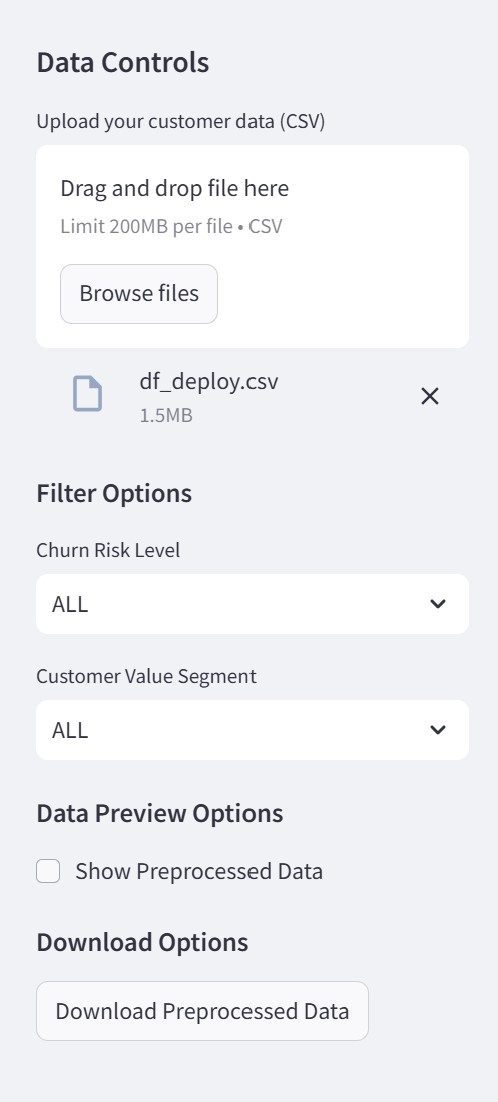}
    \caption{Sidebar data controls: CSV upload, churn-risk and value-segment filters, preprocessed-data preview toggle, and download buttons.}
    \label{fig:sidebar}
\end{figure}

Fig.~\ref{fig:landing} shows the application's landing page, which orients a first-time user with a short ``Getting Started'' checklist (upload a CSV, select target customer segments, explore insights) and a summary of the four headline features: interactive data visualization, targeted marketing recommendations, churn-risk analysis, and customer-value segmentation. Once a file is uploaded, the sidebar (Fig.~\ref{fig:sidebar}) becomes the system's primary control surface: a drag-and-drop CSV uploader; two dropdown filters for \emph{Churn Risk Level} (All / Low / Medium / High) and \emph{Customer Value Segment} (All / Low / Medium / High); a checkbox to reveal a live preview of the preprocessed, encoded dataframe; and one-click buttons to download the filtered, preprocessed data. Selecting a filter combination immediately reruns the Streamlit script and refreshes every downstream metric, chart, and recommendation without a page reload.

\subsection{Interactive Dashboards per Segment}
\begin{figure}[t]
    \centering
    \includegraphics[width=0.48\textwidth]{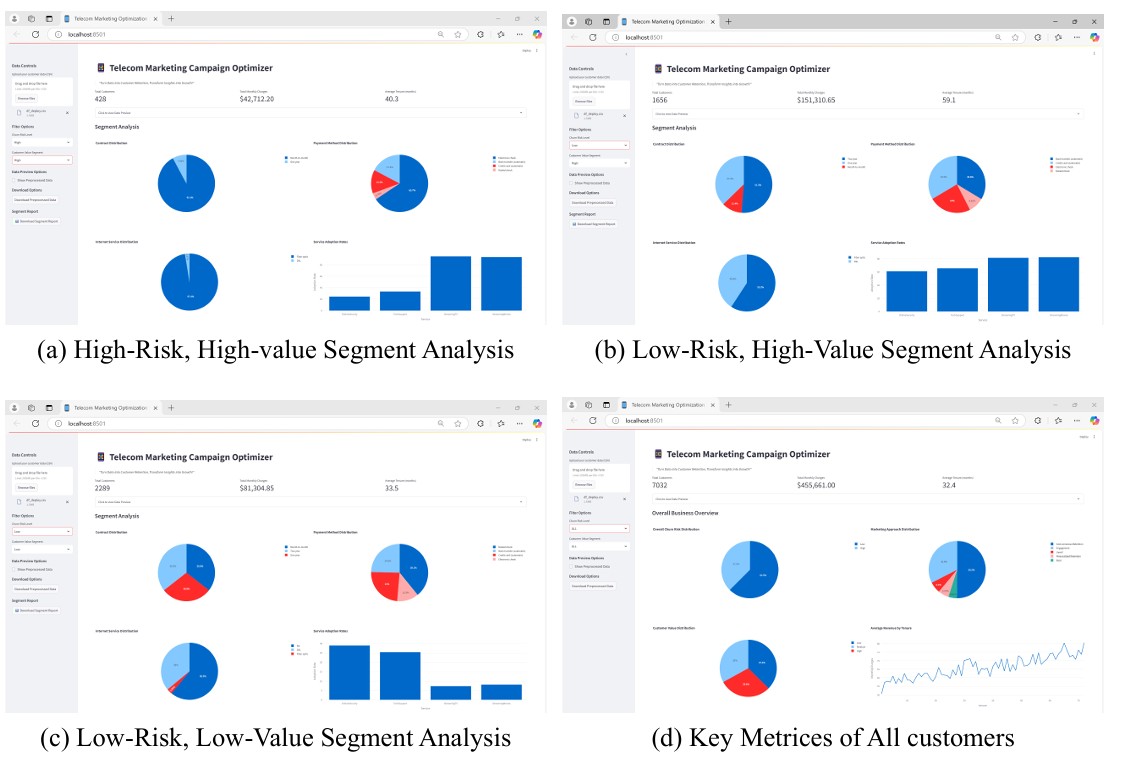}
    \caption{Key metrics and visualizations for four segment views: High-Risk/High-Value, Low-Risk/High-Value, Low-Risk/Low-Value, and All Customers.}
    \label{fig:dashboards}
\end{figure}

After filtering, the main content area renders three headline metrics (total customers, total monthly charges, average tenure in months) followed by a bank of interactive Plotly visualizations -- pie charts for churn-risk and customer-value distribution, contract/internet-service/payment-method breakdowns, and service-adoption bar charts -- all scoped to the currently selected segment. Fig.~\ref{fig:dashboards} contrasts these dashboards across four segment views: High-Risk/High-Value, Low-Risk/High-Value, Low-Risk/Low-Value, and the unfiltered ``All Customers'' overview, illustrating how the same layout automatically re-renders around whatever slice of the customer base the analyst selects.

\subsection{Automated Marketing Recommendations}
\begin{figure}[t]
    \centering
    \includegraphics[width=0.48\textwidth]{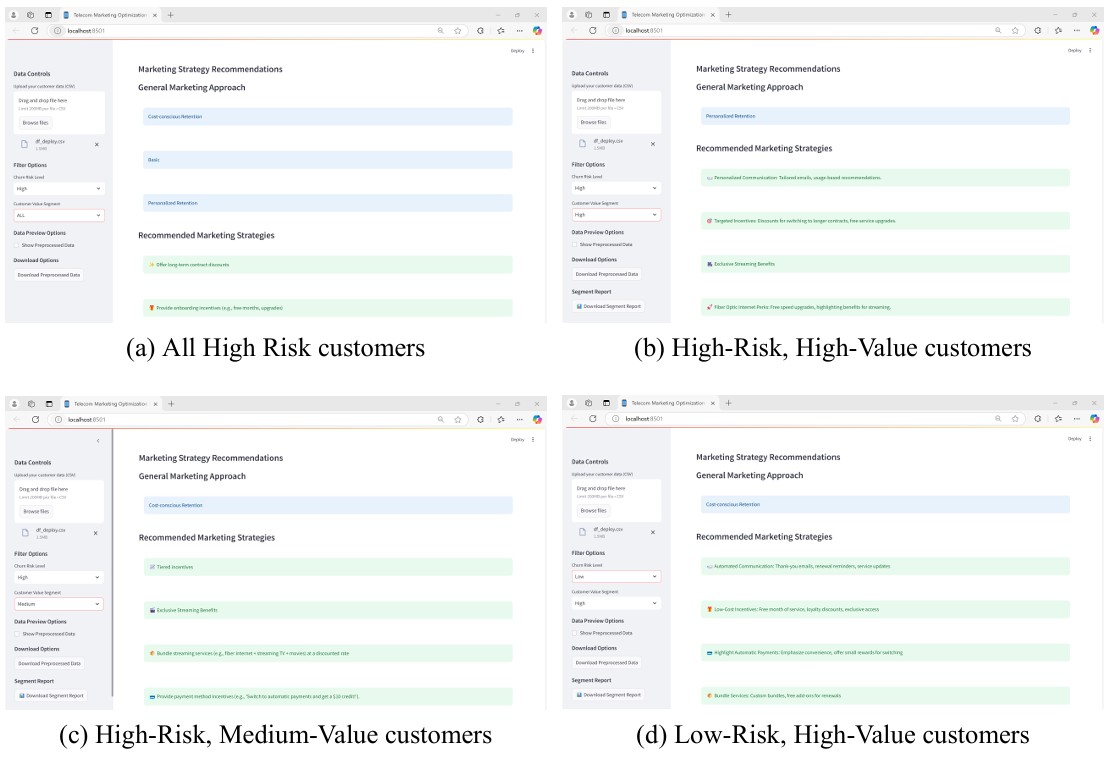}
    \caption{Automated marketing-strategy panel for four segments, each showing a distinct general approach and tactic list.}
    \label{fig:recommendations}
\end{figure}

Below the dashboards, the interface surfaces the rule-based recommendations produced by Algorithm~\ref{alg:strategy}: a short ``General Marketing Approach'' headline for the selected segment (e.g., \emph{Cost-Conscious Retention} for low-risk customers, \emph{Personalized Retention} for high-risk, high-value customers), followed by a bulleted list of concrete, segment-specific tactics -- automated communication and low-cost incentives for stable segments; free service months, exclusive content, and proactive outreach for high-risk, high-value customers; tiered incentives and service bundling for medium-value customers. Fig.~\ref{fig:recommendations} shows this panel rendered for four representative segments, demonstrating that the tactics visibly change with both the churn-risk and value axes rather than defaulting to a single generic playbook.

\subsection{Data Preview, Downloads, and Segment Reports}
\begin{figure}[t]
    \centering
    \includegraphics[width=0.48\textwidth]{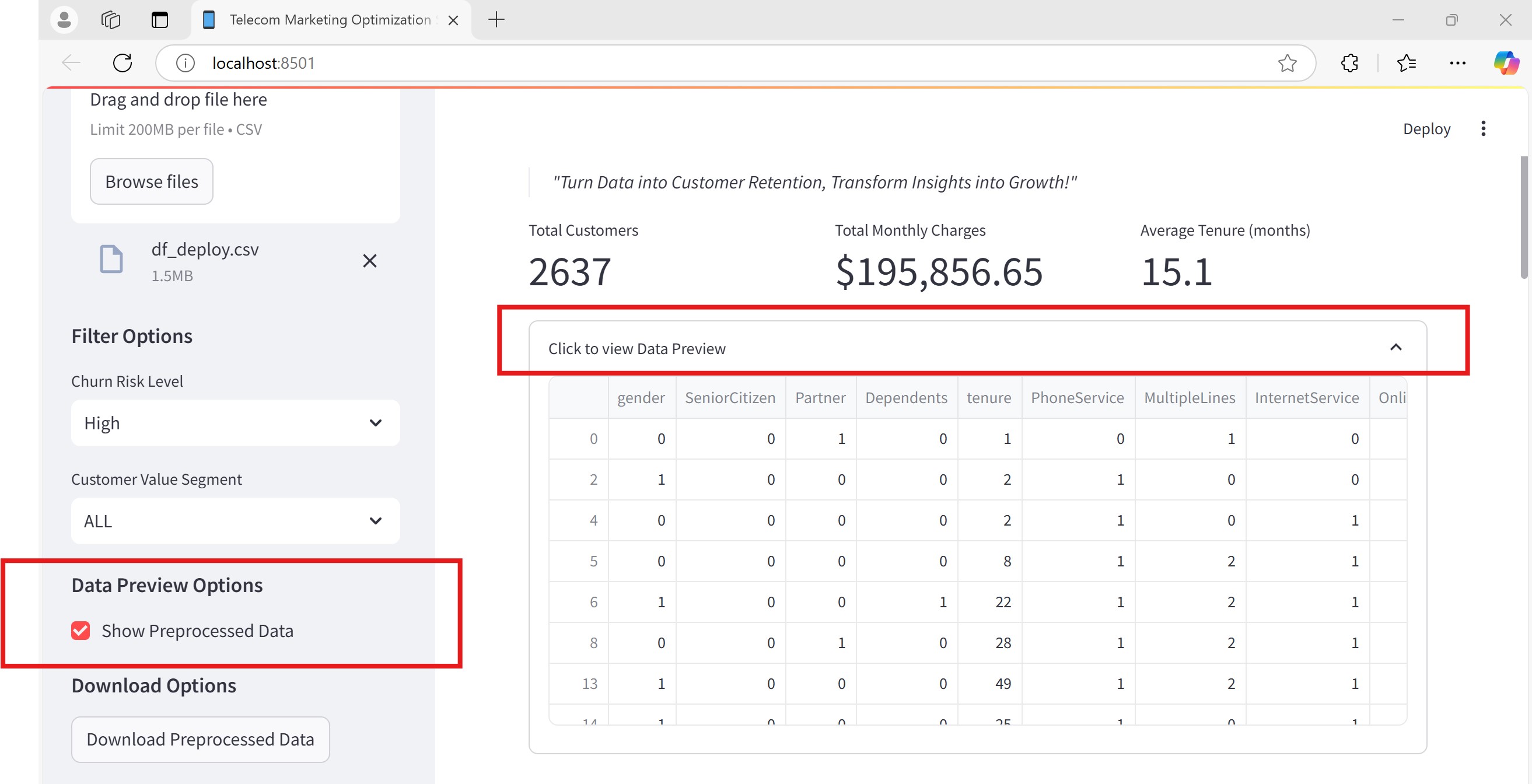}
    \caption{Expandable preview of the preprocessed, encoded dataframe, alongside the churn-risk and value-segment filter controls.}
    \label{fig:preview}
\end{figure}

\begin{figure}[t]
    \centering
    \includegraphics[width=0.42\textwidth]{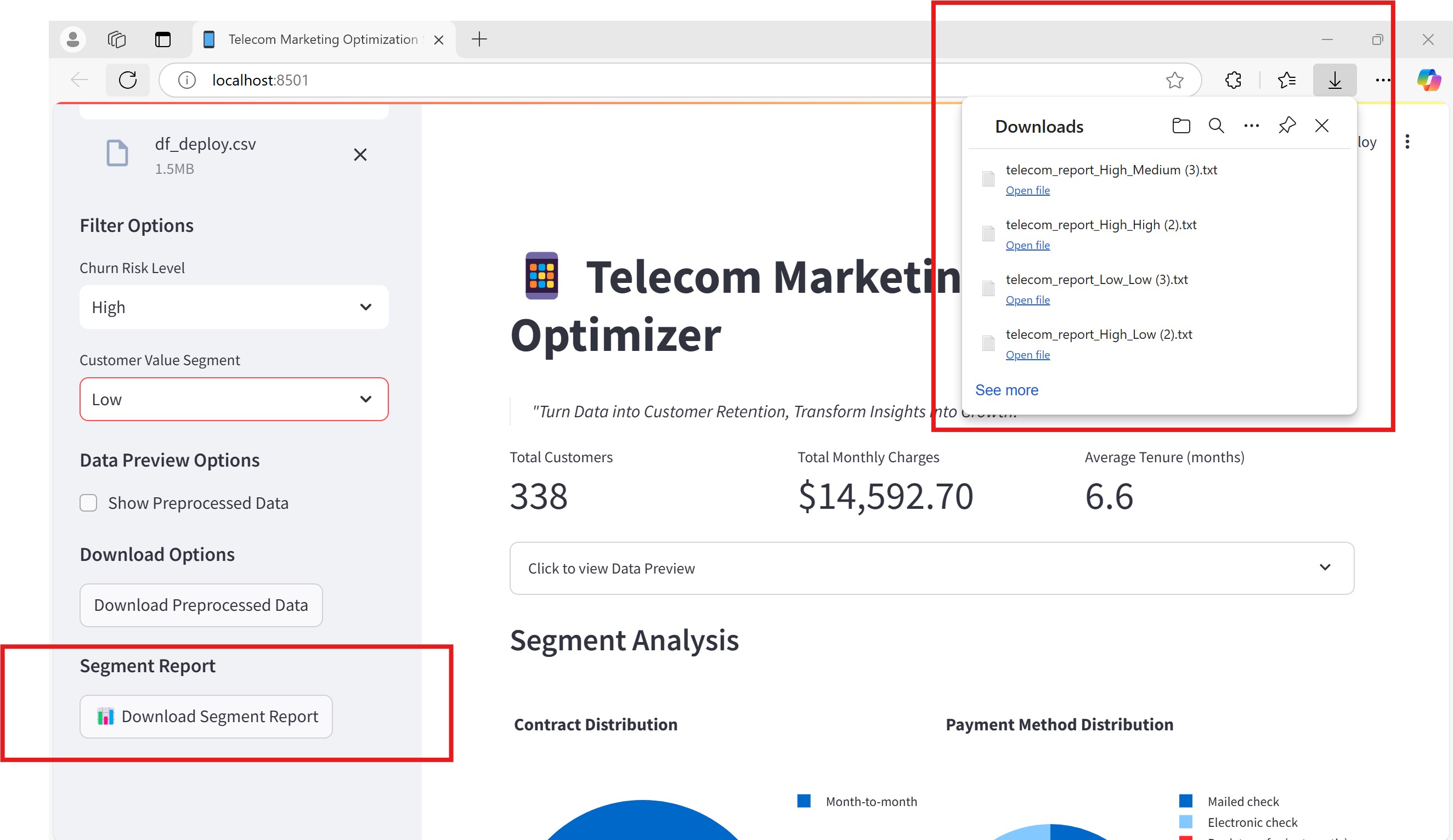}
    \caption{One-click segment-report generation and download from the sidebar.}
    \label{fig:reportbtn}
\end{figure}

\begin{figure}[t]
    \centering
    \includegraphics[width=0.42\textwidth]{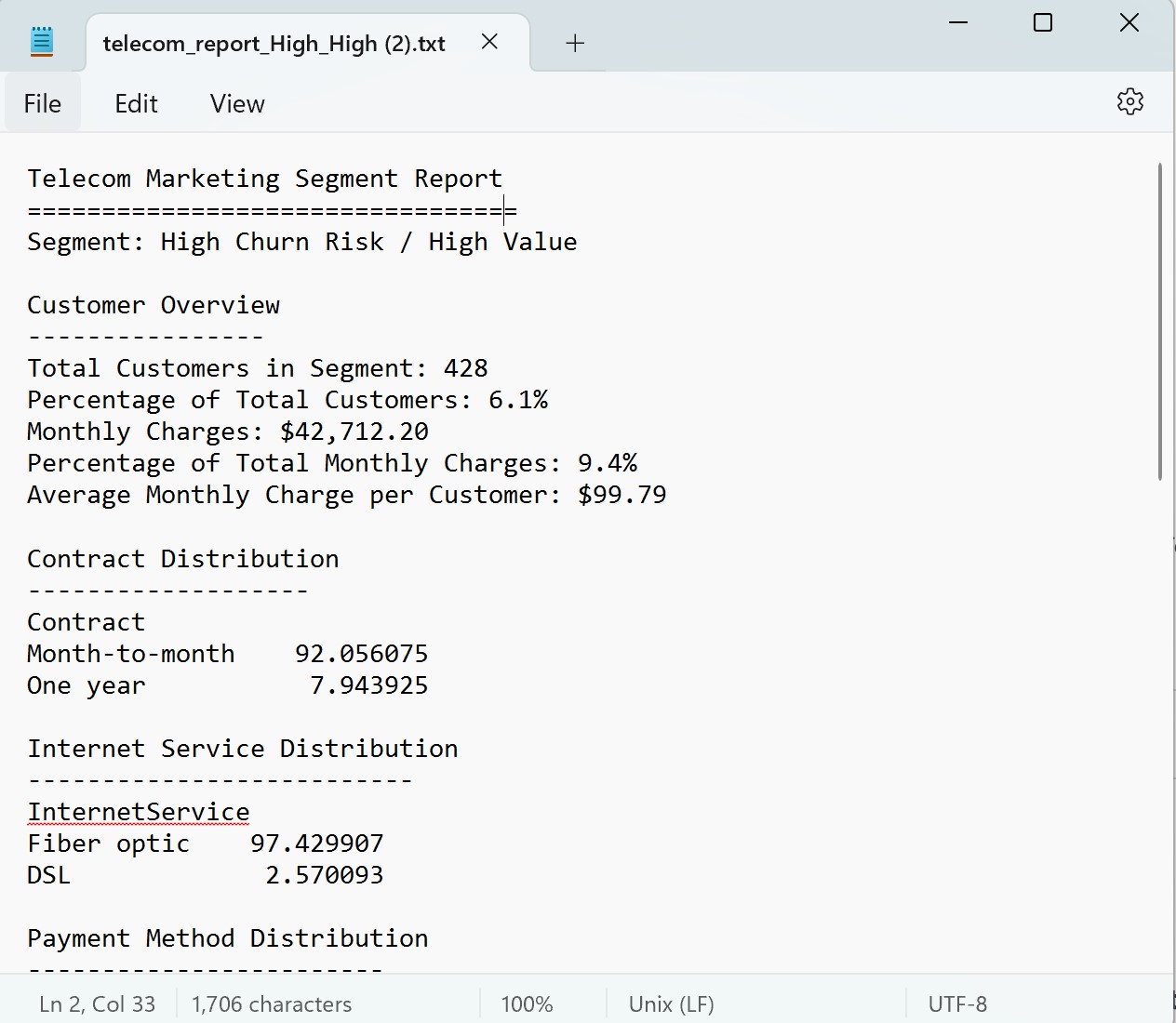}
    \caption{Sample auto-generated segment report (High-Risk/High-Value), showing customer count, revenue share, and contract/service distributions}
    \label{fig:samplereport}
\end{figure}

Two further capabilities close the loop between analysis and offline use. First, toggling ``Show Preprocessed Data'' in the sidebar expands a scrollable table preview of the encoded dataframe actually consumed by the model (Fig.~\ref{fig:preview}), letting analysts sanity-check the transformation before trusting the recommendations built on top of it; the same data can be downloaded as CSV for use in Excel or a separate BI tool. Second, a dedicated ``Download Segment Report'' button (Fig.~\ref{fig:reportbtn}) generates a templated, human-readable \texttt{.txt} report for the currently selected segment via the \texttt{generate\_segment\_report} routine described in Algorithm~2 of the accompanying thesis -- reporting the segment's customer count and percentage of the base, its share of total monthly charges, its contract/internet-service/payment-method distribution, and its adoption rate for each add-on service. Fig.~\ref{fig:samplereport} shows a representative report for the High-Risk/High-Value segment: 428 customers (6.1\% of the base), \$42{,}712.20 in monthly charges (9.4\% of total revenue), 92.1\% on month-to-month contracts, and 97.4\% on fiber-optic internet -- the exact figures underlying Table~\ref{tab:highvalue}, now packaged as a stand-alone artifact a marketing team can attach to a campaign brief without touching the underlying notebook.

\subsection{Interactive Features Summary}
In sum, the deployed dashboard provides: (1) CSV upload and dynamic filtering by churn-risk level and value segment; (2) real-time key-metric computation (customer counts, total monthly charges, average tenure); (3) interactive Plotly visualizations (pie charts for churn/value/service distributions, bar charts for service adoption, line charts for revenue trends); (4) automated, rule-based marketing-strategy recommendations per Algorithm~\ref{alg:strategy}; and (5) one-click download of preprocessed data (CSV) and detailed per-segment text reports. Every element above is generated directly from the CatBoost model and K-Means segmentation of Section IV -- there is no manual hand-off between the ML pipeline and what the marketing team ultimately sees.

\section{Experimental Results}

\subsection{Churn Prediction Model Performance}
Table~\ref{tab:positive_metrics} reports positive-class (churner) metrics and Table~\ref{tab:weighted_metrics} reports weighted-average metrics for all three tuned models on the held-out test set, after threshold optimization (Algorithm~\ref{alg:threshold}).

\begin{table}[!t]
\centering
\caption{Positive-Class (Churn) Metrics on the Test Set}
\label{tab:positive_metrics}
\resizebox{\linewidth}{!}{%
\begin{tabular}{|l|c|c|c|c|c|c|}
\hline
\textbf{Model} & \textbf{Acc.} & \textbf{Prec.} & \textbf{Rec.} & \textbf{F1} & \textbf{PR-AUC} & \textbf{ROC-AUC} \\
\hline
XGBoost & 0.7633 & 0.5386 & 0.7647 & 0.6320 & 0.6481 & 0.8408 \\
\hline
LightGBM & 0.7612 & 0.5351 & 0.7754 & 0.6332 & 0.6478 & 0.8397 \\
\hline
\textbf{CatBoost} & \textbf{0.7768} & \textbf{0.5612} & 0.7353 & \textbf{0.6366} & \textbf{0.6553} & 0.8403 \\
\hline
\end{tabular}
}
\end{table}
 
\begin{table}[!t]
\centering
\caption{Weighted-Average Performance Metrics on the Test Set}
\label{tab:weighted_metrics}
\resizebox{\linewidth}{!}{%
\begin{tabular}{|l|c|c|c|c|c|c|}
\hline
\textbf{Model} & \textbf{Acc.} & \textbf{Prec.} & \textbf{Rec.} & \textbf{F1} & \textbf{PR-AUC} & \textbf{ROC-AUC} \\
\hline
XGBoost & 0.7633 & 0.8036 & 0.7633 & 0.7741 & 0.6481 & 0.8408 \\
\hline
LightGBM & 0.7612 & 0.8051 & 0.7612 & 0.7725 & 0.6478 & 0.8397 \\
\hline
\textbf{CatBoost} & \textbf{0.7768} & 0.8041 & \textbf{0.7768} & \textbf{0.7852} & \textbf{0.6553} & 0.8403 \\
\hline
\end{tabular}
}
\end{table}

\textbf{CatBoost was selected as the deployment model}, balancing precision (56.12\%) against recall (73.53\%) with the highest F1-score (0.6366) and PR-AUC (0.6553) -- the metric most relevant under class imbalance -- while also benefiting operationally from native categorical-feature handling and 77.68\% overall accuracy. LightGBM achieved marginally higher recall (77.54\%) but at a precision cost (53.51\%), making it less suitable where false-positive retention spend must be controlled.

\begin{figure}[t]
    \centering
    \includegraphics[width=0.42\textwidth]{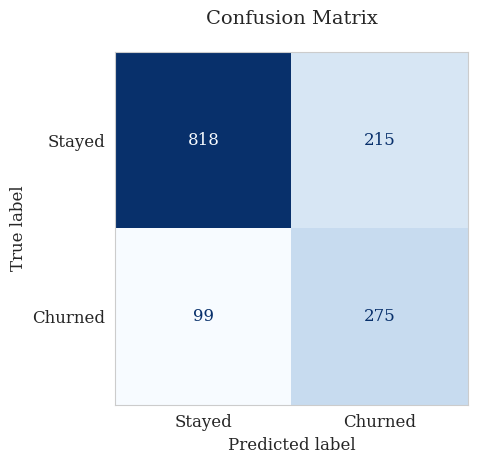}
    \caption{CatBoost confusion matrix after threshold optimization: 275 true positives, 215 false positives, 99 false negatives, 818 true negatives.}
    \label{fig:confusion}
\end{figure}

As shown in Fig.~\ref{fig:confusion}, the optimized threshold yields 275 true positives (73.53\% recall) and only 99 false negatives, confirming effective class-imbalance mitigation while limiting unnecessary retention spend to 215 false positives (56.12\% precision).

\begin{figure}[t]
    \centering
    \includegraphics[width=0.45\textwidth]{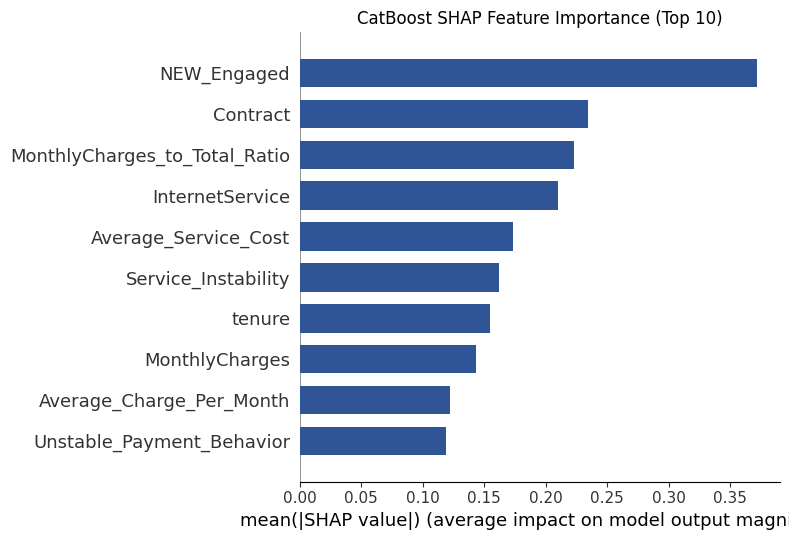}
    \caption{SHAP feature importance for the deployed CatBoost model.}
    \label{fig:shap}
\end{figure}

SHAP analysis (Fig.~\ref{fig:shap}) confirms that \texttt{Contract} type, \texttt{MonthlyCharges}, and \texttt{tenure} are the strongest churn predictors, while the engineered features \texttt{MonthlyCharges\_to\_Total\_Ratio}, \texttt{Service\_Instability}, and \texttt{Unstable\_Payment\_Behavior} rank among the top ten drivers -- empirically validating the domain-specific feature-engineering effort of Section IV-D. Customers with disproportionately high short-term spend relative to total charges show markedly higher churn propensity.

\begin{figure}[t]
    \centering
    \includegraphics[width=0.4\textwidth]{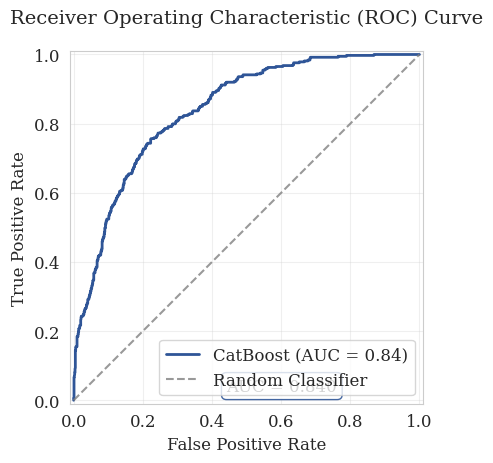}
    \caption{ROC curve for the deployed CatBoost model (AUC = 0.8403).}
    \label{fig:roc}
\end{figure}

\begin{figure}[t]
    \centering
    \includegraphics[width=0.4\textwidth]{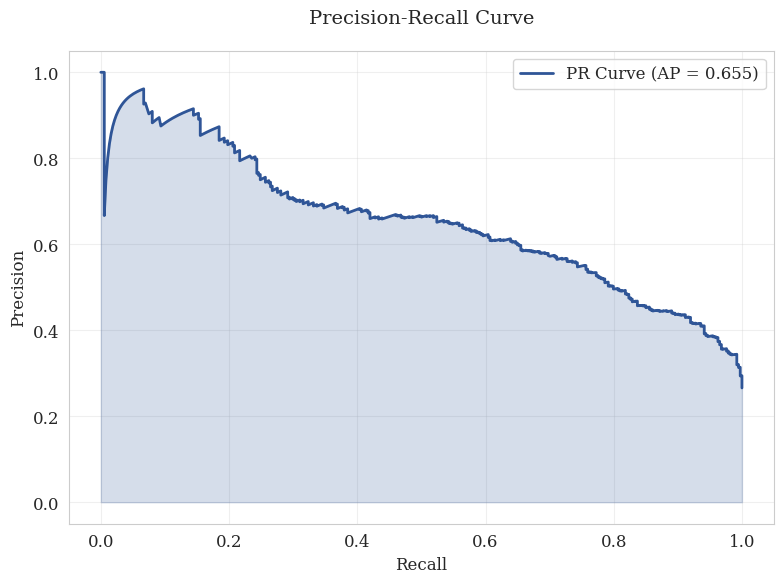}
    \caption{Precision-Recall curve for CatBoost (AP = 0.655).}
    \label{fig:pr}
\end{figure}

The ROC curve (Fig.~\ref{fig:roc}, AUC = 0.8403) confirms strong class separability, while the higher PR baseline (Fig.~\ref{fig:pr}, PR-AUC = 0.6553) demonstrates robustness specifically under class imbalance. The steep precision rise at moderate recall indicates the model's ability to prioritize high-confidence churn predictions -- precisely the behavior required for cost-efficient targeted marketing.

\subsection{Customer Segmentation Results}
\begin{figure}[t]
    \centering
    \includegraphics[width=0.45\textwidth]{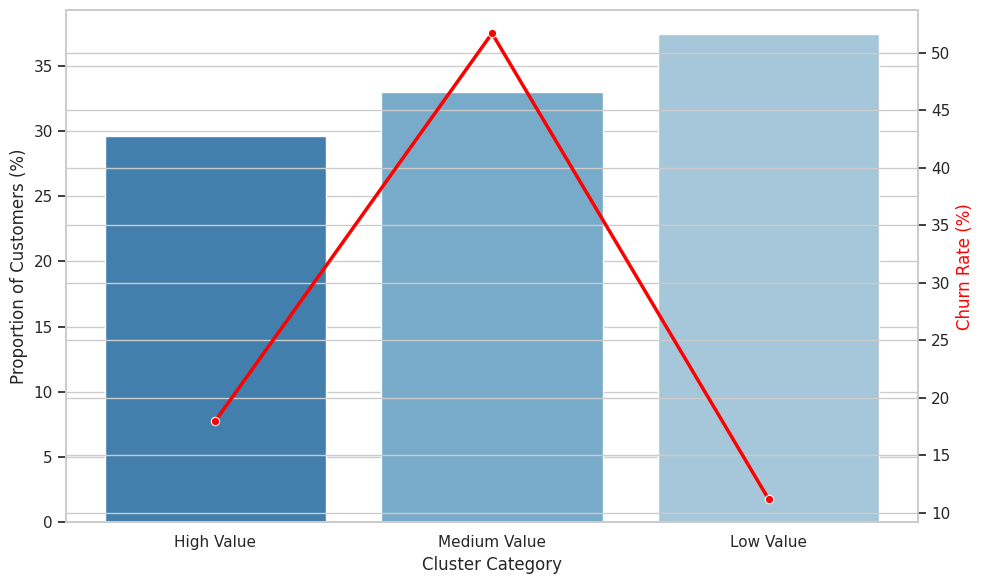}
    \caption{Proportion of customers and churn rate per value segment.}
    \label{fig:segmentation}
\end{figure}

K-Means segmentation (Section~\ref{sec:segmentation}) partitioned the customer base into High-Value (29.6\% of customers, 18.04\% churn rate), Medium-Value (33.0\%, 51.7\% churn rate), and Low-Value (37.4\%, 11.2\% churn rate) segments (Fig.~\ref{fig:segmentation}).

Notably, the Medium-Value segment exhibits by far the highest churn rate (51.7\%), despite contributing a moderate revenue share -- identifying it as the segment where retention spend has the greatest volume impact, while the smaller High-Risk/High-Value segment (6.1\% of customers, Table~\ref{tab:highvalue}) represents the highest \emph{per-customer} financial exposure.

\subsection{Marketing Strategy Optimization and ROI Outcomes}
Applying the segment-specific strategies of Section IV-F, retention campaigns targeted at high-risk, high-value customers produced the highest marginal ROI by preserving substantial future revenue, while upsell campaigns toward low-risk, high-value customers produced consistent revenue growth at low campaign cost (Table~\ref{tab:roi_scenarios}). Generalized, non-segmented retention campaigns applied uniformly to low-value customers showed diminishing returns, confirming that segmentation-based prioritization -- not churn prediction alone -- is what drives ROI efficiency.

\section{Discussion}

\subsection{Key Insights}
Combining churn prediction with value segmentation and financial evaluation produces substantially more actionable insight than any component alone. A key empirical finding is that high-value customers with only \emph{moderate} churn risk can generate greater long-term financial losses than low-value customers with \emph{high} churn probability -- reinforcing the importance of probability-based, value-aware decision-making rather than binary churn labels. The superior performance of CatBoost further confirms the practical advantage of native categorical handling on telecom-style tabular data with numerous nominal fields (contract type, internet service, payment method).

\subsection{Addressing Research Gaps}
This work addresses the three gaps identified in Section II: (1) churn prediction is embedded directly within a segmentation-and-marketing pipeline rather than treated as a stand-alone classifier; (2) segmentation explicitly incorporates churn-risk probability alongside value metrics, rather than being computed independently; and (3) the framework is operationalized as a deployed, interactive system (Section V) rather than remaining a purely offline analysis.

\subsection{Limitations}
\label{sec:limitations}
Several limitations should be considered when interpreting these results. The IBM Telco dataset is a single static snapshot lacking temporal/seasonal dynamics or external market signals (competitor pricing, macroeconomic shifts). CLV estimates rely on simplified linear-churn and uniform-ARPU assumptions (Section~\ref{sec:roi}); real deployments should recalibrate \eqref{eq:lifetime}--\eqref{eq:roi} with actual ARPU and campaign-cost data. ROI outcomes reported here are theoretical, simulated scenarios rather than live A/B-tested campaign results. Finally, the Streamlit interface currently supports only fixed visualization types and single-user sessions, and does not yet integrate multi-user role-based access or external CRM/network data streams.

\section{Conclusion and Future Work}

This paper presented a complete, reproducible ML-based marketing optimization framework for telecom churn management, integrating gradient-boosting churn prediction, two-dimensional value/risk customer segmentation, tailored marketing-strategy design, and a transparent ROI/CLV framework, all operationalized in an interactive Streamlit application. Among three tuned gradient-boosting models, \textbf{CatBoost} achieved the best balanced performance (77.68\% accuracy, F1 = 0.6366, PR-AUC = 0.6553), and SHAP analysis confirmed that both raw features (contract type, monthly charges, tenure) and engineered features (payment stability, service instability ratios) meaningfully drive predictions. Cross-tabulating churn risk with K-Means-derived value segments revealed that Medium-Value customers carry disproportionate churn risk (51.7\%) while High-Risk/High-Value customers, though only 6.1\% of the base, represent the highest per-customer financial exposure -- distinct problems requiring distinct, and now automated, marketing responses.

Future work will (1) incorporate real-time behavioral streams (support logs, network-quality metrics, social signals) for temporal/survival-based churn modeling; (2) explore reinforcement learning for continuously self-optimizing campaign allocation; (3) extend the framework to other subscription-based industries (streaming, SaaS, utilities) where churn dynamics are structurally similar; and (4) add multi-user, role-based access to the Streamlit deployment to support collaborative use by larger marketing organizations.

\section*{Code and Data Availability}
The complete implementation -- churn-model notebooks, the integrated segmentation/marketing pipeline, and the Streamlit deployment package -- is available at the authors' GitHub repository (contact us if you need any clarification). The IBM Telco Customer Churn dataset is publicly available on Kaggle \cite{ref28b}.

\bibliographystyle{IEEEtran}

\end{document}